\documentclass{article}
\usepackage[final]{colm2026_conference}  %
\usepackage{microtype,booktabs,graphicx,amsmath,amssymb,array}  %
\usepackage{hyperref,url}
\usepackage{pgfplots}\pgfplotsset{compat=1.18}  %
\definecolor{darkblue}{rgb}{0, 0, 0.5}
\hypersetup{colorlinks=true, citecolor=darkblue, linkcolor=darkblue, urlcolor=darkblue}
\makeatletter
\def\@listi{\leftmargin\leftmargini \topsep 2pt plus 1pt minus 1pt \parsep 0pt \itemsep 1pt}
\let\@listI\@listi
\makeatother

\title{Who Maintains Agent Skills?\\A Longitudinal Study of Human-Governed, AI-Assisted Skill Maintenance}
\author{Chen Shen \& Estevam Hruschka \\
Megagon Labs \\
\texttt{\{chen\_s,estevam\}@megagon.ai} \\}

\begin{document}
\maketitle
\lhead{Accepted at the COLM 2026 Workshop on Lifelong Agents}
\vspace{-12pt}

\begin{abstract}
\vspace{-4pt}
Lifelong LLM agents increasingly rely on external skill artifacts as one element for preserving and reusing capabilities over time.
These skills (usually portable Markdown files such as \texttt{SKILL.md}) describe when and how to apply a capability and must be corrected, expanded, and consolidated as tools and usage patterns shift over deployment.
Recent work seeks to automate skill curation, but it largely evaluates against automated baselines and treats human maintenance as an unmeasured bottleneck.
We study that missing process directly.
We mine the full commit histories of five public AI-skill repositories, a purposive sample of AI-tooling organizations, covering 873 commits, 143 skill files, and 254 substantive post-creation edits from October 2025 to June 2026.
We code each edit with pre-registered governance, operation, and trigger-evidence codebooks.
Three findings emerge.
First, every substantive edit is authored or merged through a named human account, while 62\% carry an AI co-author trailer, with large repository-level variation.
Second, these edits are genuine curation: an audited sample shows that most change skill content, and the coded operations are dominated by additions and corrections.
Third, a pre-registered rule-likeness axis fails its reliability gate; reliably coding rule-likeness from commit artifacts remains an open measurement problem.
We release the corpus, codebooks, mining scripts, and a replay protocol for automated skill curators.
For self-evolving agents, public skill maintenance currently looks less like an autonomous pipeline than a human-governed, AI-assisted loop that future curators must measure against and operate within.
\end{abstract}

\section{Introduction}\label{sec:intro}
\vspace{-4pt}
Agent skills are becoming an important element in the operational memory of LLM agents: small, portable Markdown packages that tell an agent when and how to apply a capability and can play a key role in the self-evolving and self-improving capabilities of the agents.
Unlike benchmark prompts or one-off demonstrations, skills live under version control.
They are corrected when tools change, expanded when maintainers discover missing cases, reorganized when instructions become unwieldy, and occasionally retired when capabilities become obsolete.
Recent 2025--2026 work automates this curation: systems learn to edit skill documents from rollouts \citep{yangSkillOpt2026}, distill trajectories into skills \citep{niTrace2Skill2026}, and train curators that evolve an external skill repository \citep{ouyangSkillOS2026}.
These systems are routinely evaluated against automated or memory-based baselines; manual maintenance is treated mainly as a scalability concern, ``demand[ing] huge human expertise and cannot scale'' \citep{ouyangSkillOS2026}, and is, to our reading, mostly unmeasured.
Even the strongest evidence that humans matter concerns an endpoint: a recent benchmark reports human-authored skills at 74.5\% task success, while no automated method averages above 31.1\% across the evaluated models \citep{zhongSkillLearnBench2026}, but says nothing about the human process that produced them.
In this paper, we measure that process.

This leaves a prerequisite empirical question: who maintains agent skills in public repositories, and what does that maintenance consist of?
Closest to us, a concurrent study of agent \emph{context} files (\texttt{AGENTS.md}/\texttt{README}-style) finds they ``evolve like configuration code through frequent, small additions'' \citep{agentReadmes2025}.
But it studies a different artifact and characterizes maintenance only at that coarse level, without a coded operation taxonomy, commit-authorship governance, or failure-trigger evidence.
A marketplace analysis of skills is cross-sectional \citep{lingClaudeSkillsAnalysis2026}, and decision-history methods construct their histories synthetically \citep{liSkillHone2026}.
We are not aware of a longitudinal study of public Markdown \emph{skill-file} maintenance histories and their curation operations.

To characterize this process, we mine git histories directly.
From a purposive sample of five public AI-skill repositories (873 commits at pinned clones), we track 143 \texttt{SKILL.md} files through 631 commit-file history records, isolate 254 substantive post-creation maintenance edits, and code each under three pre-registered codebooks.
Our contributions are:
\vspace{-4pt}
\begin{itemize}\itemsep1pt\parsep0pt
\item \textbf{A released corpus and protocol} for mining skill-file maintenance: commit-level records, three codebooks, and reproducible scripts (\S\ref{sec:method}).\footnote{\url{https://github.com/megagonlabs/who-maintains-agent-skills}} The corpus can support replay evaluations of automated skill-curation systems (protocol in Appendix~\ref{app:testbed}).
\item \textbf{The governance finding} (\S\ref{sec:rq3}): trailer-visible AI co-authorship under human repository control. Of the substantive edits, 62\% carry an AI co-author trailer, 38\% are AI-trailer-absent, and every edit is authored or merged through a named human account, which is all we mean by human-governed. The mix differs sharply by organization, a disclosure-and-merge-culture split as much as an AI-usage one.
\item \textbf{A coded account of maintenance} (\S\ref{sec:rq1}, \S\ref{sec:rq3}): operation types, observable failure evidence, and descriptive size/cadence trajectories (\S\ref{sec:rq2}).
\item \textbf{A negative result} (\S\ref{sec:rq1}): a pre-registered ``rule-likeness'' axis failed its reliability gate; we report this rather than the unreliable distribution, and use it to motivate future work (\S\ref{sec:discussion}).
\end{itemize}
The study is descriptive: the five repositories are a purposive, not representative, sample of AI-tooling organizations, and our claims are limited to what their maintenance histories contain.

\section{Related Work}\label{sec:related}
\vspace{-4pt}
\textbf{Automated skill curation} builds on lifelong-agent skill libraries that agents grow autonomously, from \citet{wangVoyager2024}'s ever-growing library of executable skills to reflective successors that distill experience into reusable insights \citep{shinnReflexion2023,zhaoExpeL2024}.
Recent work edits or evolves skill text from experience through text-space optimizers \citep{yangSkillOpt2026}, trajectory distillation \citep{niTrace2Skill2026}, RL-trained curators of an external repository \citep{ouyangSkillOS2026}, and fully autonomous self-evolving skill-library systems \citep{linMUSEAutoskill2026,shenSkillFoundry2026,yuSkillAdaptor2026}.
Benchmarks measure whether the resulting skills help \citep{zhongSkillLearnBench2026,liSkillsBench2026,hanSWESkillsBench2026}.
None studies the human maintenance process, and \citet{ouyangSkillOS2026} runs no manual-curation arm.

\textbf{Empirical mining of evolving text artifacts} is our closest methodological area: studies of context-file evolution and adoption \citep{agentReadmes2025,mohsenimofidiContextEng2026}, agentic-tool configuration datasets \citep{agentConfigDataset2026}, prompt evolution in repositories \citep{tafreshipourPromptingWild2025}, AI-IDE rule files \citep{caiAIIDERules2026}, and CI-workflow evolution \citep{rostamiGHActionsEvolution2026} all follow the same workflow: open-code a sample into a typed change taxonomy, then scale it.
The prompt- and context/rule-file studies report the additive-dominant signature we also find \citep{tafreshipourPromptingWild2025,caiAIIDERules2026}, while workflow edits skew modification-heavy \citep{rostamiGHActionsEvolution2026}.
\citet{tafreshipourPromptingWild2025} also show only 21.9\% of prompt changes in commit messages, echoing our sparse trigger evidence.
The closest concurrent work is a registered report that plans to mine agent context files (\texttt{CLAUDE.md}/\texttt{AGENTS.md}) and derive a maintenance-grounded change taxonomy \citep{voriaACFMaintenance2026}; it excludes skill files and does not attribute who makes each edit.
We instead study executable \texttt{SKILL.md} skills, apply a pre-registered operation taxonomy, and focus on the governance question of who authors or merges each edit.
Those studies mine $10^3$--$10^4$ changes, far more than our 254 coded edits; what our smaller corpus adds is the \texttt{SKILL.md}-specific measurement none of them applies to skill files: commit-authorship governance and coded curation operations.

Our \textbf{governance signal} (the \texttt{Co-Authored-By} commit trailer as an AI-authorship marker) follows established repository-mining practice \citep{mineArchTrailers2026}.

\textbf{Structured skill/experience memory} includes typed skill graphs co-trained with a policy \citep{liSkillGraphRL2026}, graph-structured experience memory \citep{fengExpGraph2026}, and reflective semantic/episodic memory \citep{hassellSemanticEpisodicMemory2026}; \textbf{runtime correction} compiles user corrections into enforcement hooks \citep{zhouTraceEnforcement2026}.
Both operate on internal stores or runtime artifacts; none mines longitudinal skill-file histories or writes generalized rules back into skill files (Table~\ref{tab:diff}).

\begin{table}[t]\centering\small
\resizebox{\columnwidth}{!}{%
\begin{tabular}{@{}lccc@{}}
\toprule
 & Long. & Object & Curation \\
\midrule
SkillOpt / SkillOS / Trace2Skill \citep{yangSkillOpt2026,ouyangSkillOS2026,niTrace2Skill2026} & \checkmark & skill files & automated only \\
SkillLearnBench / SWE-Skills-Bench \citep{zhongSkillLearnBench2026,hanSWESkillsBench2026} & -- & skill files & endpoint only \\
Agent READMEs \citep{agentReadmes2025} & \checkmark & context files & coarse, uncoded \\
Claude-skills marketplace \citep{lingClaudeSkillsAnalysis2026} & -- & skill files & -- \\
SkillHone \citep{liSkillHone2026} & \checkmark & skill files & synthetic \\
SkillGraph / ExpGraph \citep{liSkillGraphRL2026,fengExpGraph2026} & \checkmark & internal graph & -- \\
TRACE \citep{zhouTraceEnforcement2026} & \checkmark & runtime rules & user corrections \\
\textbf{This work} & \checkmark & \texttt{SKILL.md} edit histories & \textbf{coded + measured} \\
\bottomrule
\end{tabular}}
\caption{Where this study sits. To our knowledge, prior work has not mined longitudinal \emph{maintenance histories} of human-curated Markdown skill files with coded operations and governance. Long.\ = longitudinal; Object = the artifact studied; Curation = coded human-curation analysis.}\label{tab:diff}
\end{table}

\section{Corpus and Method}\label{sec:method}
\vspace{-4pt}
\textbf{Corpus.} To study public skill maintenance where it is observable, we purposively selected five public repositories that maintain agent skills as first-class Markdown files and exhibit sustained commit activity: \texttt{getsentry/skills}, \texttt{trailofbits/skills}, \texttt{obra/superpowers}, \texttt{cloudflare/skills}, and \texttt{anthropics/skills}.
These AI-tooling organizations use their own agent products, spanning application monitoring (Sentry), security research (Trail of Bits), a single-maintainer personal skill collection (obra), internet infrastructure (Cloudflare), and an AI lab (Anthropic).
Before any governance or operation coding, we froze the repository set, pinned each clone to a fixed commit, and did not run a systematic census of candidate repositories.
Maintenance-depth quantities are therefore likely to resemble upper bounds for broader public skill repositories, and governance shares are descriptive and organization-dependent rather than population base rates, with the direction of any broader bias left unestimated (external validity is revisited in \S\ref{sec:discussion}).
Across the pinned clones, the corpus spans 2025-10 to 2026-06: 873 commits across the five repositories, 143 \texttt{SKILL.md} files yielding 631 commit-file history records, of which 254 are substantive post-creation edits (defined below).
Table~\ref{tab:corpus} gives the per-repository pins and counts.

\begin{table}[t]\centering\
{\scriptsize
\begin{tabular}{@{}lrrrc@{}}
\toprule
repository & \texttt{SKILL.md} & subst.\ edits & edit-days & span \\
\midrule
getsentry/skills    & 28 & 81 & 40 & 2026-01--2026-06 \\
trailofbits/skills  & 74 & 78 & 22 & 2026-01--2026-06 \\
obra/superpowers    & 14 & 62 & 26 & 2025-10--2026-05 \\
anthropics/skills   & 18 & 19 & 11 & 2025-11--2026-06 \\
cloudflare/skills   &  9 & 14 & 10 & 2026-02--2026-04 \\
\midrule
\textbf{total}      & \textbf{143} & \textbf{254} & & \\
\bottomrule
\end{tabular}
}
\caption{The purposive corpus: per-repository skill-file count, substantive edits, distinct edit-days, and span (months). Pinned commits are listed in the reproducibility note.}\label{tab:corpus}
\end{table}
\vspace{-4pt}

\textbf{Unit and substantive-edit definition.} The unit of analysis is one \texttt{SKILL.md} file tracked through renames (\texttt{git log -{}-follow}).
A \emph{substantive post-creation edit} is any commit after the file's creation that changes at least five lines of it and is neither bot-authored nor trivial.
This yields 254 substantive edits.
Because \S\ref{sec:rq3} reports no agent-only substantive commit, we checked whether the bot filter could have created that observation.
The filter removes only 13 bot-authored \texttt{SKILL.md} commits, all routed through the platform web-merge identity, and only 8 of them would otherwise qualify as substantive ($\approx$3\% of the would-be-substantive set).
The ``no agent-only substantive commit'' observation in \S\ref{sec:rq3} is therefore not an artifact of this exclusion.
A pre-registered sensitivity battery (Appendix~\ref{app:sensitivity}) recomputes the count of deeply maintained skills ($\ge 3$ substantive edits) across thresholds and mass-refactor exclusion.
With 27 / 22 / 16 / 15 skills across the four cells, the corpus supports a descriptive study in every cell.
A 50-edit audit sample categorized 70\% as skill-content changes (the remainder migration, dependency-version churn, or pure style), confirming the substantive-edit definition is not dominated by mechanical churn.

\textbf{Three codebooks.} To connect each edit to the study questions, we coded every substantive edit along three axes: \emph{governance}, derived deterministically from commit metadata (bot author; AI co-author trailer; human author with web-merge committer; human author and committer); \emph{curation operation}, under a pre-registered eight-operation taxonomy we constructed for skill-file maintenance that can be collapsed onto the classical corrective-versus-enhancement split used in \S\ref{sec:rq1}; and \emph{trigger evidence} on a 0--3 ladder (L3 explicit issue/failure link; L2 named concrete failure; L1 generic; L0 none).
For operation, trigger, and rule-likeness labels, a single pass of Opus 4.8 coded all 254 edits, and a second pass of the same model coded a 50-edit stratified sample in a separate session.
The coding model received the codebook, the commit subject, and the per-file diff.
Unlike the de-identified judging in \S\ref{sec:rq4}, these inputs were not identity-scrubbed (paths and diffs can reveal the repository), which we record as a limitation.
We report agreement as Cohen's $\kappa$, raw agreement corrected for chance \citep{cohen1960}, and use the 50-edit overlap to estimate coding stability rather than human--human agreement or validation against human labels.
Two passes that share a systematic bias can agree while being jointly wrong, so we treat the 8-way operation taxonomy as descriptive and rest headline claims on the two-way collapse.
That collapse is validated by a blind single-coder human recode of the 50-edit sample at $\kappa=0.72$ (8-way leaves only $\kappa=0.46$) and corroborated, as a post-submission sensitivity check, by a cross-family recode of all 254 edits (collapse $\kappa=0.65$, raw $0.83$; Appendix~\ref{app:codebooks}).
Governance is mechanical, so the dual-coding was a quality-control check: coded labels matched the deterministic value on 252 of 254 edits (99.2\%), and the two disagreements were annotation errors corrected to the metadata-derived value.
All scripts and coded data are released.

\section{What Maintenance Does}\label{sec:rq1}
\begin{figure}[t]\centering
\includegraphics[width=1.0\linewidth]{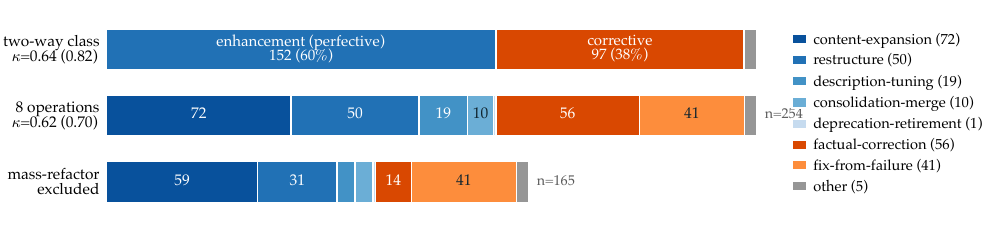}
\caption{\textbf{Operations over the 254 substantive skill-file edits.} Middle row: the eight coded operations (counts in the legend). Top row: the same edits grouped into the corrective and enhancement maintenance classes. Bottom row: the recount on the 165 edits that remain after excluding multi-file mass-refactor commits.}
\label{fig:tax}
\end{figure}

\vspace{-4pt}
With the corpus and codebooks in place, we first ask what maintainers change when they edit skill files, and the answer leans toward enhancement over repair.
Figure~\ref{fig:tax} reports the distribution of curation operations across the 254 substantive edits; content expansion (72) and factual correction (56) together account for half of all edits.
We collapse the eight operations onto the standard corrective-versus-enhancement split of classical maintenance taxonomy \citep{swanson1976dimensions,iso14764} as an interpretive grounding, not a recoding: factual-correction and fix-from-failure are corrective, and the remaining five operations are grouped as enhancement (perfective improvement).
Under this collapse the corpus is 60\% enhancement and 38\% correction, with 2\% other (Figure~\ref{fig:tax}, top row).
In practice, roughly three edits enhance a skill for every two that repair one.

\vspace{-4pt}

\textbf{Reliability and denominators.} Because the operation labels are LLM-coded, we use the reliability checks of \S\ref{sec:method} to separate headline claims from descriptive detail.
At the leaf level the labels are only moderately stable: inter-LLM-pass agreement on the eight-way taxonomy is $\kappa=0.62$, and cross-family re-codes agree less ($\kappa=0.36$/$0.40$ on 50 edits, $0.46$ over all 254).
The two-way collapse is sturdier: it raises raw inter-LLM agreement to 0.82 at a similar $\kappa$ (0.64 vs.\ 0.62), and the blind human recode validates it at $\kappa=0.72$ (Appendix~\ref{app:codebooks}), which is why headline claims rest on the collapse while leaf counts stay descriptive.
The denominator also matters: we count per skill-file edit row, so a multi-file commit contributes one row per skill it touches.
Under stricter denominators the skew only sharpens: the 50-edit audit flagged $\sim$30\% of rows as non-curation (migration, dependency churn, style), the 35 curation-confirmed audit edits give 77\%/23\% (Appendix~\ref{app:sensitivity}), and weighting commits equally gives 67\% enhancement, 31\% corrective, and 2\% other over the 187 commits, leaving the row-level 60\% the lowest of the weightings we examine.
We do not claim a complete maintenance classification: adaptive (migration) edits fall outside the codebook's scope and appear only as audit residue (8 of the 50 audited edits).

\textbf{The corrective share is partly bulk repair.} Of the factual-correction edits, 75\% come from multi-file mass-refactor commits, chiefly one \texttt{trailofbits} spec fix that contributes 25 edit-rows.
Under mass-refactor exclusion, factual correction falls from the second-largest category to the fourth, while content expansion, fixing-from-failure, and restructuring are robust (Appendix~\ref{app:sensitivity}).
Consolidation/merging (10) and deprecation (1) are rare: skills accrete and are corrected far more than they are pruned, a pattern directly relevant to the unbounded-growth concern in never-ending skill learning.

\textbf{A pre-registered rule-likeness axis was not reliable enough to report.} To ask whether human curation produces generalizable knowledge, we pre-registered a rule-likeness axis: whether an edit is ``rule-like'' (a generalizable constraint) or ``instance-bound'' (a one-off fix).
It did not pass its reliability check: inter-rater $\kappa = -0.02$, with raw agreement of only 0.56 and disagreement concentrated at the instance-bound/mixed boundary (confusion matrix, Appendix~\ref{app:codebooks}).
The negative $\kappa$ partly reflects a base-rate paradox.
The second pass labels almost everything ``rule-like'' (38/50), so observed agreement (0.56) sits just below chance (0.57) and $\kappa$ turns negative.
Prevalence-adjusted, PABAK${}=0.34$ and Gwet's AC1${}=0.44$, so the failure is as much an instrument and base-rate effect as evidence the two constructs are intrinsically inseparable.
Per our pre-registration we do not report the apparent distribution as a finding.
Still, the failure is informative: if two independent LLM passes cannot reliably separate rule-like from instance-bound edits, systems that would generalize curation rules face the same coding obstacle (\S\ref{sec:discussion}).

\section{How Skills Evolve}\label{sec:rq2}
\vspace{-4pt}
After coding what maintenance does, we next ask how skill files change over time.
Across the 120 skills with at least two size observations, trajectories are heterogeneous but lean toward growth: 32 grow by more than 10\% in resident token length, 7 shrink by more than 10\%, and 81 remain stable.
Successive commits touch a skill every 5 days at the median (488 intervals over all skill-file commits, including non-substantive ones, and the median remains 5 days when restricted to substantive edits).
Given 5--8 month right-censored windows and an uneven depth distribution, the data do not support a strong growth-or-decay narrative: maintenance is frequent and predominantly stable-to-additive, with neither broad compounding nor net decay.
This is a suggestive, loosely drawn contrast with the addition-dominated picture reported for agent context files \citep{agentReadmes2025}.\footnote{The artifacts and metrics differ: their per-commit word add/delete counts versus our net resident-token trajectories.}
Net shrinks are uncommon, but consolidation and restructuring remain part of the repertoire (\S\ref{sec:rq1}), consistent with skills accreting and being corrected far more than they are pruned. Observed corrective or failure-related edits are not defect-rate measures: they reflect what is visible in maintenance histories and may equally indicate active detection, documentation, and repair of issues.
Per-repository edit counts and the deeply maintained-skill distribution are in Appendix~\ref{app:sensitivity}. The cadence and size-change distributions are in Figure~\ref{fig:rq2dyn}.

\textbf{Maintenance concentrates on the skill body.} We decompose each \texttt{SKILL.md} into six components with a deterministic parser (frontmatter; the name/description ``router''; instructions; examples; code; references) and attribute the changed lines in every substantive edit.
The body dominates: 85\% of edits touch instructions and 56\% touch embedded code, while the router (38\%), frontmatter (15\%), examples (11\%), and standalone references (4\%) are edited far less often.
\section{Who Maintains, and Why}\label{sec:rq3}
\vspace{-4pt}
\textbf{Governance: trailer-visible AI co-authorship under human repository control.} Of 254 substantive maintenance edits, 158 (62\%) carry an AI co-author trailer in the commit body, while the remaining 96 (38\%) are AI-trailer-absent (60 human-authored-and-committed, 36 human-authored\footnote{We consider accounts named as human-owned to define ``human-authored'' actions.} and web-merged). Counting commits rather than skill-file edit rows leaves the split essentially unchanged (61\%, 114 of 187 commits). Put operationally, three of every five substantive edits carry disclosed AI co-authorship, each released through a human account.
Within the non-bot substantive set, every edit is authored or merged by a named human, and we observe no agent-only substantive commit, a claim about commit attribution rather than verified authorship. Bot commits are excluded by definition, and only 8 would-be-substantive bot commits exist (\S\ref{sec:method}), so the exclusion does not manufacture this observation.
Governance is therefore a spectrum: the 62\% AI-trailered edits are human--AI collaborations, and the rest show no trailer-visible AI.

\textbf{No size or scope separation.} We detect no size or scope separation between AI-trailered and AI-trailer-absent edits: medians of 15 versus 21 changed lines (Mann--Whitney $p=0.20$) and 4 versus 6 files ($p=0.31$).
The governance signal tracks who authors or merges an edit and carries no detectable information about what the edit changes.

We next ask whether editor identity predicts which part of a skill changes.
No component's governance split (AI-trailered vs.\ AI-trailer-absent) survives our pre-registered controls (Holm-corrected Fisher, mass-refactor exclusion, and repository-stratified Cochran--Mantel--Haenszel).
The raw cross-tab misleads: frontmatter edits look $90\%$ AI-trailered against a $62\%$ base rate, but the signal is entirely a multi-file mass-refactor artifact, collapsing to the base rate (on six edits) once fan-out commits are excluded.
Together with the size result above, we therefore detect no governance-by-component specialization: across these repositories, who authors or merges an edit predicts neither how large it is nor which part of the skill it changes.
Because the battery is low-powered and multiply tested, this null shows only the absence of a detectable division of labor, not strict uniformity (Appendix~\ref{app:component}).

\begin{figure}[t]\centering
\includegraphics[width=0.92\linewidth]{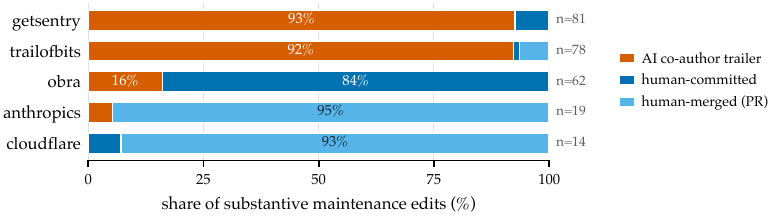}
\caption{\textbf{Human-governed maintenance; trailer-visible AI co-authorship varies by organization.}
Repository shares by governance class from commit metadata (counts in Table~\ref{tab:gov}); every substantive edit is human-authored or human-merged. Classes indicate trailer-visible co-authorship, not measured AI use; squash workflows can drop trailers (\S\ref{sec:rq3}).These repository-level shares should not be interpreted as measures of organization-wide AI adoption, human contribution, or engineering practice.}
\label{fig:gov}
\end{figure}

\textbf{The regime is organization-dependent.} The regime varies sharply by organization (Figure~\ref{fig:gov}, Table~\ref{tab:gov}): a single-maintainer project is almost entirely human-authored; two organizations route nearly all maintenance through human-merged pull requests with no AI trailer; two others co-author the large majority of edits with an AI tool.
The pooled 62\% therefore summarizes a bimodal regime, not a population base rate. Because pooling trailer signals across heterogeneous repositories can reverse the apparent conclusion \citep{yuSimpsonCoauthorship2026}, we report governance per organization.
The observed mix can reflect several factors that we cannot disentangle, including actual AI involvement, trailer-generation practices, merge/squash workflows, and repository-specific attribution conventions.”

Across all five repositories, every edit is still human-authored or human-merged. Automated curation systems that operate on these public artifacts would need to fit that repository process.

\textbf{Construct validity.} Four constructs are distinct here.
True AI use we do not measure.
Trailer-visible AI co-authorship is the deterministic signal we report (a \texttt{Co-Authored-By} trailer naming an AI tool, \citealp{mineArchTrailers2026}); it undercounts undisclosed use \citep{robbesMiningAgentActivity2026,kraishanAIAttributionParadox2025} and can overcount auto-injected trailers.
Human authorship-or-merge metadata establishes that a named human account authored or released each edit.
Verified human review is what commit metadata cannot establish: ``human-merged'' denotes the pull-request web-merge flow, not confirmed review approval.
Our claims concern the middle two, and ``human-governed'' in this paper means exactly that human account-holders control authorship and release.
A concrete anchor for the possible undercount: \texttt{anthropics/skills} reads as almost entirely human-merged with no AI trailer (Table~\ref{tab:gov}), yet Anthropic reports that, as of May 2026, more than 80\% of the code merged into its production codebase is authored by Claude under an engineer-directed review model \citep{anthropicRecursiveSelfImprovement2026}.
Because that repository is documentation rather than production code, this example anchors the plausible magnitude of undercount rather than a rate for our corpus.

\vspace{-4pt}
\begin{table}[t]\centering
{\scriptsize
\begin{tabular}{@{}lrccc@{}}
\toprule
repository & $n$ & AI & HC & HM \\
\midrule
getsentry/skills & 81 & 75 & 6 & -- \\
trailofbits/skills & 78 & 72 & 1 & 5 \\
obra/superpowers & 62 & 10 & 52 & -- \\
anthropics/skills & 19 & 1 & -- & 18 \\
cloudflare/skills & 14 & -- & 1 & 13 \\
\midrule
\textbf{total} & \textbf{254} & \textbf{158} & \textbf{60} & \textbf{36} \\
\bottomrule
\end{tabular}
}
\caption{Governance by repository (deterministic from commit metadata). AI = AI co-author trailer present; HC = human-committed; HM = human-merged via pull request, with no trailer-visible AI co-authorship.}\label{tab:gov}
\end{table}

\textbf{Trigger evidence.} We quantify observable evidence linking an edit to a usage failure, not a causal claim.
Of 254 edits, 61 (24\%) carry concrete failure evidence (L2 or L3); 12 cite an explicit issue or describe a reported failure (L3).
Most edits (193, L1) are generic maintenance whose connection to a specific failure is not stated in the record.
We therefore reserve ``failure-triggered'' for the L2/L3 minority and report the denominator.
No edit fell to L0 in the primary pass, so its realized ladder is L1--L3.
The $\kappa = 0.59$ inter-pass reliability is borderline, the disagreement is directional (the second pass reads more failure evidence), and a post-submission cross-family recode of all 254 edits under the identical codebook (\texttt{gpt-5.5}) reads concrete evidence into 63\% of edits ($\kappa=0.18$ against the primary pass), almost entirely by promoting generic-cue (L1) edits to L2 while confirming 59 of the 61 primary-pass L2/L3 edits.
The concrete-evidence share is thus instrument-dependent: 24\% is the stricter reading, not a stable quantity; we treat the trigger counts as descriptive only (a calibration caveat for codebook reusers), drawing no structural claim from them.
\vspace{-4pt}
\section{Does Maintenance Help? A Bounded Transfer-Task Test}\label{sec:rq4}
\vspace{-4pt}
Beyond describing maintenance, we asked whether it improves a skill.
An earlier pilot hinted that the maintained version outperforms the earliest in-window one, but it had four weaknesses: the solver saw only the \texttt{SKILL.md} file rather than the full skill package, the exchange was limited to a single turn, the judge came from the same provider as the solver, and the design had roughly 20\% statistical power (Appendix~\ref{app:rq4-pilot}).
We pre-registered and conducted a replication that fixes each of these. The full design, statistics, and per-skill numbers appear in Appendix~\ref{app:rq4}.
The study uses 13 public skills with $\ge 6$ substantive edits across four repositories, each paired with 11 transfer tasks (143 total), attempted by a solver from a different model family than the judges (\texttt{gpt-5.4-mini}) under \{\texttt{no-skill}, \texttt{v\_old}, \texttt{v\_new}\}; the solver sees the full skill package through on-demand file reads, an equalized multi-turn protocol lets a constrained user-simulator answer clarifying questions from the brief alone, and a sealed panel of blind judges scores de-identified, fingerprint-scrubbed, randomized triads.
The inference unit is the skill ($n=13$), with pre-registered power of 0.77 to 0.90 for the pilot-sized effect at a fixed sample size.
At the skill level the effect on judged quality is indistinguishable from zero: the mean \texttt{v\_new}${}-{}$\texttt{v\_old} difference is $-0.09$ on the 1--5 scale, with a 95\% confidence interval of $[-0.28,+0.10]$. Neither a sign test ($p=0.58$) nor a Wilcoxon test ($p=0.55$) rejects the null, and 5 of 13 skills favor \texttt{v\_new}.
This result survives a length-controlled re-analysis and a weaker-solver re-run under \texttt{gpt-4o-mini} (Appendix~\ref{app:rq4}).
We read this as no evidence for a pilot-sized benefit of maintenance under this harness, not as evidence that maintenance is inert.
The interval still admits small positive effects, and cross-family judge agreement was only moderate (Spearman 0.51; single-measures ICC(2,1)${}={}$0.52 on the 100-triad overlap, below the pre-registered 0.6 gate, Appendix~\ref{app:rq4}), so a small true effect could be attenuated by measurement noise.
Whether maintenance helps a downstream solver is thus design-contingent: the pilot's positive trend does not persist under the powered design that controls package exposure, interaction channel, judge family, and statistical power.
This is a bounded null on author-and-model-constructed transfer tasks.
Our descriptive contribution does not rest on this result.

\begin{figure}[t]\centering
\includegraphics[width=1.0\linewidth]{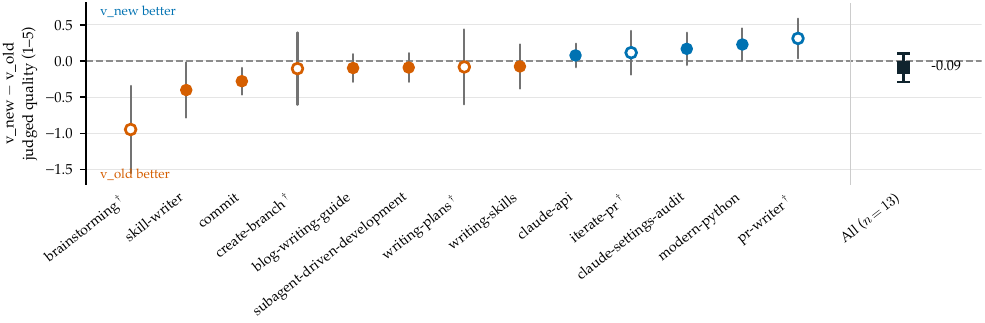}
\caption{\textbf{Maintenance moves individual skills both ways, but not on average.}
Maintained-minus-earliest judged-quality difference (1--5 rubric). Points show skill means over 11 tasks $\times$ 3 replicates; whiskers show descriptive task-level 95\% CIs; $\dagger$ marks deep-history skills; square marks the pre-registered skill-level aggregate. Per-skill differences are specific to this constructed transfer-task harness and should not be interpreted as measures of intrinsic skill quality, maintainer quality, or effectiveness in the repositories' native workflows.}
\label{fig:rq4}
\end{figure}

\textbf{Why the per-skill effects split.} Two extremes explain the split (illustrative only).
\texttt{brainstorming} (most negative, $-0.95$) and \texttt{pr-writer} (most positive, $+0.31$) both roughly doubled to tripled in length under maintenance, so size is not the discriminator.
\texttt{brainstorming}'s edits added structured elicitation steps (a pre-design gate, a checklist, a process diagram). On a one-shot transfer task this steered the solver toward process management and suppressed design content (judged coverage $2.06\!\to\!1.33$, both skilled versions below the skill-free baseline).
\texttt{pr-writer}'s edits instead added output constraints (rewrite-don't-append, prose over headings, bans on placeholder references), which transferred to the judged artifact (non-redundancy $3.45\!\to\!4.58$, the largest single-dimension gain).
Across the $13$ skills these two regimes cancel ($5$ positive, $8$ negative): the pilot's apparent benefit averages out once a heterogeneous skill set is scored on transfer tasks, so a powered aggregate reads null while individual skills still move both ways.

\textbf{Relation to in-the-wild skill evaluations.} Our powered null aligns with empirical skill-benefit studies that find benefits fragile.
In-the-wild benchmarks find skill benefit to be setting-dependent but refinement-recoverable \citep{liuSkillsInWild2026}, and library growth can degrade agents through skill shadowing \citep{songSkillShadowing2026}. Both vary deployment or library, whereas we hold the harness fixed, vary only the maintained version, and observe neither benefit nor shadowing-style loss.
Continual-skill benchmarks ask whether machine-generated or continually evolved skills help \citep{zhongSkillLearnBench2026,liSkillHone2026}. We ask whether human maintenance helps a downstream solver, finding no pilot-sized transfer benefit, so ``does maintenance help'' is itself a measurement question.
Native repository workflows or external skill benchmarks could differ, and we release the powered harness to test that \citep{kimBiGGenBench2025}.

\section{Discussion: Toward Rule-Generalizing Curation}\label{sec:discussion}
\vspace{-4pt}
\textbf{From field observations to a curator.} Two results constrain curator design.
First, maintenance is a standing human--AI loop that grows and corrects skills far more than it prunes them: the budget pressure anticipated by never-ending skill learning appears here as accretive, human-governed editing.
Second, the rule-likeness reliability failure shows that ``does this edit encode a generalizable rule?'' is hard to operationalize reliably from commit artifacts, as two independent LLM coding passes did not reach acceptable agreement.
Together, these results suggest a curator that records traceable edit rationales and consolidates them into maintainer-approved generalized guidance, rather than rewriting skill text from individual instances.
This is closer to never-ending learning over a typed knowledge base \citep{mitchellNeverEndingLearning2015} than one-shot text optimization, and complements critique-distilling reflective memory without such structure \citep{hassellSemanticEpisodicMemory2026}.
Similarly, ``LLM wiki'' envisions an agent-maintained interlinked Markdown knowledge base at near-zero human cost \citep{karpathyLLMWiki2026}, and WiNELL trains agents on historical edits to propose never-ending Wikipedia updates for human review \citep{reddyWiNELL2026}. Our measurement shows the observed loop still runs through human-owned repository processes.
Designing such a curator remains future work. The measured regime and released corpus instead yield design hypotheses.
\textbf{Design implications for autonomous skill-curators.} These hypotheses are grounded in the observed maintenance regime, not validated designs.
\vspace{-4pt}
\begin{enumerate}\itemsep0.5pt\parsep0pt
\item \emph{Budget explicitly for pruning.} Consolidation and deprecation are only $4.3\%$ of edits (11/254), so a curator may need an explicit consolidation/retirement budget.
\item \emph{Keep curation under human-owned release or merge control.} Every substantive edit was authored or merged through a named-human repository process, so autonomous curators should surface proposed edits at that boundary.
\item \emph{Do not rely solely on failure-provenance to supervise.} The concrete-failure-evidence share is instrument-dependent (24\% under our primary pass, 63\% under a cross-family recode), the trigger ladder is only borderline-reliable even within one family, and rule-likeness failed its gate (Sections~\ref{sec:rq1},~\ref{sec:rq3}); failure provenance cannot yet be measured stably enough to serve as a curator's primary supervision target.
\item \emph{Triage by operation type before size.} Operation type is our highest-agreement coded axis (\S\ref{sec:rq1}), while governance predicts neither edit size nor scope (\S\ref{sec:rq3}), so a curator should route first by what an edit does rather than by edit size.
\item \emph{Model governance per repository, not as one global mix.} The $62\%$ AI-trailer prevalence is an observed, two-sided proxy (it can under- and over-count true AI involvement), and the split is bimodal across repositories, so a curator deployed across organizations should model local disclosure and merge cultures rather than assume a single global human--AI mix.
\end{enumerate}
The released corpus makes these hypotheses testable: a candidate curator sees a prior skill version and context, proposes the next edit, and is scored against the human edit (Appendix~\ref{app:testbed}). Building a baseline remains future work.

\section{Conclusion}\label{sec:conclusion}
\vspace{-4pt}
Skill maintenance in these public repositories is frequent, human-governed in the specific sense that every edit is authored or merged through a named human account, and often visible in commit trailers as AI-assisted.
Prior work on automated curation has largely left this maintenance process unmeasured.
We release a commit-level corpus, three codebooks, and reproducible scripts for studying this process, and we report a pre-registered rule-likeness failure that clarifies the open problem for rule-generalizing curators.
For lifelong, self-evolving agents, these repositories suggest that curation systems should be evaluated against human-governed maintenance loops, not only against fully automated skill-evolution pipelines.

\bibliography{refs}
\bibliographystyle{colm2026_conference}

\appendix
\section*{Limitations}
The sample is purposive (five AI-tooling organizations) and selected on sustained maintenance, so maintenance-depth quantities are upper-bound-like, while governance shares are descriptive and organization-dependent, not population base rates. Also, because inclusion requires observable public histories, the study has an inherent transparency-selection effect: organizations and maintainers that expose richer histories make their corrections, AI-attribution signals, and maintenance activity more measurable than less-open counterparts. These observations should therefore not be interpreted as evidence of lower engineering quality, greater AI dependence, or weaker maintenance practices.
Trajectories are right-censored over 5--8 months.
The operation and trigger codebooks are LLM-coded: the corrective/enhancement collapse is human-validated at $\kappa=0.72$ with a single coder over 50 edits; the 8-way leaves were human-checked but agreed only moderately ($\kappa=0.46$); and the trigger axis has no human check; a full cross-family recode reproduces neither its labels nor its distribution ($\kappa=0.18$; 63\% vs.\ 24\% L2+L3), so trigger counts are primary-pass-specific.
The governance signal is observed \texttt{Co-Authored-By} trailer prevalence, which can both under- and over-count true AI involvement.
Trigger evidence is bounded by what commit records make observable, and ``human-merged'' denotes the pull-request web-merge flow, not verified review approval.

\section*{Ethics and reproducibility}
All data are public repository histories; no human subjects are involved.
We note nonetheless that public availability is not consent to be aggregated: our governance signal derives from named human committer/author and AI-tool identities.
We therefore report only organization-level aggregates (Table~\ref{tab:gov}), not individual authorship profiles, and our released records expose only what the public commit metadata already contains (mined commit metadata and our own coding labels, with no repository file content redistributed); we add no inferred attributes and apply no de-anonymization beyond the trailer and merge-committer fields git records.
The released records do join these public identity fields to our curation labels; this adds no non-public identity data, and we report analyses only at the organization level, noting that for a single-maintainer repository organization-level aggregation is effectively individual-level.
\textbf{Released artifacts} (the resource contribution), available at:\footnote{\url{https://github.com/megagonlabs/who-maintains-agent-skills}} (i) commit-level records, 631 \texttt{SKILL.md} history records over 143 files, with the 254 substantive edits labeled under all three codebooks; (ii) a codebook datasheet with label definitions and deterministic/judgment decision rules; (iii) the mining, component-attribution, LLM operation-coding, and offline-analysis scripts, with a reproduction script; and (iv) pointers to the five pinned public source repositories (a clone script pinning each commit) with their own licenses.
Source repositories are pinned at fixed commits (getsentry \texttt{b39c7c4}, trailofbits \texttt{c070b9b}, obra \texttt{6fd4507}, anthropics \texttt{5754626}, cloudflare \texttt{12520fd}) and carry their own licenses (\texttt{trailofbits} is CC-BY-SA); we redistribute no repository file content, only derived metadata and our labels, so the release is license-compatible.
The labeled-edit numbers (Sections~\ref{sec:rq1}--\ref{sec:rq3} and the appendices) recompute offline from the released records and annotation passes with no API key; the corpus-level commit counts (e.g.\ the 873 commits) are reproducible by re-cloning the five pinned repositories and re-running the deterministic mining script.
The operation taxonomy is reproducible up to LLM coding noise with any annotator model, while the paper's exact operation distribution and $\kappa$ recompute offline from the released passes; the maintenance-benefit analysis (\S\ref{sec:rq4}) regenerates from the archived solver and judge logs, which are frozen and not bit-reproducible but included in the release.

\section{Codebooks and Reliability}\label{app:codebooks}
The operation codebook assigns one label per edit by \emph{dominant maintenance purpose}, with tie-breaks toward the more specific label.
Two axes are kept deliberately separate: \texttt{fix-from-failure} is reserved for edits whose dominant purpose is repairing an \emph{observed execution failure} (else the same content change is \texttt{factual-correction}); it is thus an operation-with-observed-failure-provenance category, not a pure edit-mechanism one, and trigger evidence is coded separately for evidence strength (we therefore do not treat operation and trigger as statistically independent axes); \texttt{factual-correction} changes truth conditions whereas \texttt{description-tuning} changes only the router/wording; and \texttt{restructure}, \texttt{consolidation-merge}, and \texttt{deprecation-retirement} are distinguished by whether content is reorganized, merged, or removed.
The five \texttt{other} edits are non-curation residue (e.g.\ licensing/boilerplate).
The released codebooks datasheet carries the full definitions, precedence rules, and per-label examples; Table~\ref{tab:opcodebook} summarizes the eight operations with an illustrative edit and the key boundary for each.

\begin{table}[h]\centering\footnotesize
\begin{tabular}{@{}l p{0.28\textwidth} p{0.22\textwidth} p{0.19\textwidth}@{}}
\toprule
operation & definition & illustrative edit & distinguished from \\
\midrule
\texttt{content-expansion} & adds new guidance, sections, or examples (substance) & adds a ``common pitfalls'' subsection & description-tuning (wording only) \\
\texttt{factual-correction} & fixes an incorrect statement, value, or API & corrects a flag name or version string & fix-from-failure (no failure cue) \\
\texttt{restructure} & reorganizes existing content & splits one section into three & content-expansion (adds substance) \\
\texttt{fix-from-failure} & responds to an observed usage failure & adds a stop condition after a reported loop & factual-correction (no failure cue) \\
\texttt{description-tuning} & rewords the name / description / trigger text & tightens the ``use when'' line & content-expansion (adds substance) \\
\texttt{consolidation-merge} & merges or dedupes content & folds two overlapping steps into one & restructure (no dedupe) \\
\texttt{deprecation-retirement} & marks obsolete / removes a capability & adds ``deprecated: use X instead'' & restructure (keeps content) \\
\texttt{other} & non-curation residue & a license or boilerplate line & the seven substantive labels \\
\bottomrule
\end{tabular}
\caption{The eight curation operations with an illustrative edit and the boundary distinction that most often decides the label. The released datasheet carries full per-label examples and precedence rules.}\label{tab:opcodebook}
\end{table}

\begin{table}[h]\centering\footnotesize
\begin{tabular}{@{}l >{\raggedright\arraybackslash}p{0.24\textwidth} >{\raggedright\arraybackslash}p{0.34\textwidth} r@{}}
\toprule
coded operation & repository / skill & commit subject (verbatim) & lines \\
\midrule
\texttt{content-expansion} & getsentry / blog-writing-guide & \begin{tabular}[t]{@{}l@{}}add bad/good examples to\\AI patterns section\end{tabular} & 34 \\
\texttt{factual-correction} & trailofbits / solana-vulnerability-scanner & \begin{tabular}[t]{@{}l@{}}correct duplicate section\\numbering\end{tabular} & 15 \\
\texttt{fix-from-failure} & getsentry / create-branch & \begin{tabular}[t]{@{}l@{}}Simplify Step 5 and fix\\unconditional stash pop\end{tabular} & 61 \\
\texttt{restructure} & getsentry / agents-md & \begin{tabular}[t]{@{}l@{}}Improve agent\\guidance skills\end{tabular} & 147 \\
\bottomrule
\end{tabular}
\caption{Four real coded edits from the corpus, grounding the codebook in concrete cases. Subjects are quoted from the commits (conventional-commit prefixes and PR numbers elided); lines = changed lines in the named skill file.}\label{tab:realexamples}
\end{table}

\textbf{Rule-likeness confusion.} The $3\times3$ confusion matrix (pass A rows, pass B columns) is: rule-like $\to$ (28, 1, 7); instance-bound $\to$ (6, 0, 4); mixed-unclear $\to$ (4, 0, 0).
Disagreement concentrates off the rule-like diagonal.
Because pass B assigns ``rule-like'' to 38/50 edits, the low $\kappa$ is prevalence-driven, a base-rate paradox: $p_o=0.56$, $p_e=0.57$, $\kappa=-0.02$, while the prevalence-adjusted PABAK${}=0.34$ and Gwet's AC1${}=0.44$.
This is the basis for the downgrade in \S\ref{sec:rq1}.
A later cross-family recode of the same axis as a single binary item, where a reusable rule is present only if a specific added line states one, reaches $\kappa=0.43$ (raw agreement $0.72$) between two different-family models; this suggests the failure was largely the abstract three-way operationalization rather than the underlying construct, and we still do not report the rule-likeness distribution as a finding.

\textbf{Uncertainty.} The same-family inter-pass $\kappa$ values are point estimates on the 50-edit overlap; we report no confidence intervals for the inter-LLM pass agreements, and at $n{\approx}50$ a 95\% interval on $\kappa=0.62$ is wide enough to straddle the conventional 0.70 threshold, so the $0.62$ vs.\ $0.59$ ordering should be read as indistinguishable.

\textbf{Cross-family check.} To test whether the operation reliability is an artifact of coding with one model family, we re-coded the same 50-edit overlap with two different-family models (\texttt{gpt-4o} and \texttt{gpt-5.5}) under the identical codebook.
The 8-way operation agreement with the Claude pass falls to $\kappa=0.36$ and $0.40$ (below the same-family inter-pass $0.62$), so the fine-grained leaf labels carry agreement specific to the coding model family.
The two-way corrective-versus-enhancement collapse that our headline result uses, however, reaches $\kappa=0.51$ and $0.61$ (raw agreement $0.76$ and $0.80$), below but close to the same-family collapse $\kappa=0.64$.
Cross-family agreement is thus substantially better for the collapse than for the eight leaves, though still only moderate; it corroborates the enhancement-versus-correction finding against a same-family-bias confound and supports resting claims on the collapse rather than the eight leaf labels.
A post-submission recode of all 254 edits with \texttt{gpt-5.5} under a bundled three-axis prompt corroborates this picture at scale: 8-way $\kappa=0.46$, collapse $\kappa=0.65$ (raw $0.83$), with the collapse distribution moving at most 3.5 points ($60/38/2 \to 58/42/0.4$).
The trigger axis does not survive the same recode ($\kappa=0.18$; \S\ref{sec:rq3}), and the three-way rule-likeness axis remains unreliable ($\kappa=0.17$) despite near-identical marginals: the two families assign the same distribution to different edits, confirming the axis measures noise.
\textbf{Human validation.} One author then independently re-coded the 50-edit overlap blind to the LLM labels.
Human-vs-LLM agreement is \emph{moderate} on the 8-way taxonomy ($\kappa=0.46$, 95\% CI $[0.29,0.62]$, raw $0.58$; below the inter-LLM-pass $0.62$, as expected if two passes of one model share a systematic bias) but \emph{substantial} on the two-way corrective-versus-enhancement collapse our headline uses ($\kappa=0.72$, CI $[0.51,0.88]$, raw $0.86$; above the inter-LLM collapse $0.64$).
Fourteen of the 21 leaf-level disagreements stay within the same corrective/enhancement bucket (concentrated at the content-expansion/restructure and factual-correction/fix-from-failure boundaries), so they do not move the headline.
This is a small blind human check of the headline split and indicates the eight leaves are only moderately human-valid, which is why we rest claims on the collapse; it is a single coder ($n=50$), so it does not establish a human--human reliability ceiling.
\textbf{Governance proxy.} We similarly checked the AI-trailer signal: the same coder judged whether \emph{independent} (non-trailer) evidence of AI involvement was visible on 50 commits.
Clear auto-injection was rare (1 of 25 trailer-present commits, 4\%) and 2 of 25 trailer-absent commits showed independent AI evidence despite no trailer, but the majority carried no independent signal either way (31/50 \emph{unclear}).
The trailer is thus rarely a clear false positive yet usually not independently verifiable, which supports reporting the 62\% strictly as \emph{trailer-visible} AI co-authorship rather than an estimate of true AI use.
\section{Component Decomposition and Governance}\label{app:component}
We parse each \texttt{SKILL.md} version into six components by a deterministic, total-precedence rule (YAML frontmatter; \texttt{name}/\texttt{description} ``router'' keys and matching headings; instructions; examples; code fences; standalone references), and attribute each substantive edit's changed lines by diffing the file against its parent (254/254 edits attributed; the parser is released).
Edits touching each component (of 254): instructions 215 (85\%), code 142 (56\%), router 97 (38\%), frontmatter 39 (15\%), examples 28 (11\%), references 11 (4\%).
For each component we test the association between ``edit touches it'' and governance class (AI-trailered vs.\ human-only) with three pre-registered controls: a Holm-corrected Fisher exact test, the same test with multi-file mass-refactor commits excluded, and a repository-stratified Cochran--Mantel--Haenszel test.
No component survives all three.
The largest raw effect is frontmatter, which looks $90\%$ AI-trailered against the $62\%$ base rate (Fisher $p<10^{-3}$); the effect disappears under mass-refactor exclusion, where only six frontmatter edits remain ($p=1.0$), and under repository stratification ($p=0.22$).
It is a fan-out artifact of a few AI-co-authored mass refactors.
The router shows a residual human lean ($47\%$ AI-trailered) that holds under mass-refactor exclusion ($p=0.015$) but not under repository stratification ($p=0.12$), so we do not assert it.
We report no component-attribution reliability coefficient here (the parser is deterministic; a line-level agreement study against an independent recode is future work), and the null is robust to the unambiguous frontmatter-versus-body split.
\section{Sensitivity Analyses}\label{app:sensitivity}
Deeply maintained skills ($\ge 3$ substantive edits) by threshold and mass-refactor exclusion: $\ge\!5$ lines incl.\ mass = 27; $\ge\!10$ incl.\ = 22; $\ge\!5$ excl.\ = 16; $\ge\!10$ excl.\ = 15.
\textbf{Operation distribution under mass-refactor exclusion} (89 of 254 edits are multi-file mass-refactor; 165 remain): content-expansion 59, fix-from-failure 41, restructure 31, factual-correction 14, description-tuning 7, consolidation-merge 7, other 5, deprecation 1.
Content expansion, fix-from-failure, and restructuring are robust; factual-correction drops from rank 2 to rank 4 (75\% mass-refactor, 25 of its rows from one \texttt{trailofbits} spec fix).
Per-repository edit counts and the 50-edit audit sample (35 curation / 8 migration / 5 dependency-version / 2 docs-style).
Re-running the corrective/enhancement collapse on the second-pass labels of the 35 curation-confirmed edits gives 77\% enhancement / 23\% corrective, versus 70\% / 26\% / 4\% other on all 50, so restricting to confirmed curation sharpens rather than weakens the enhancement skew reported in \S\ref{sec:rq1}.

\begin{figure}[h]\centering
\includegraphics[width=0.98\linewidth]{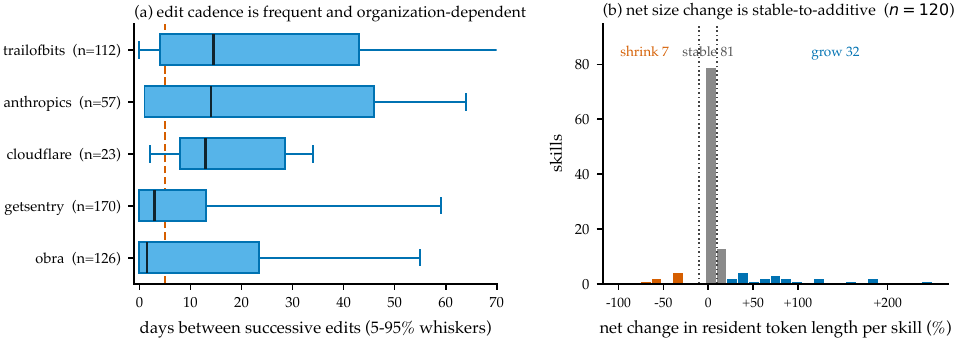}
\caption{\textbf{Maintenance dynamics (\S\ref{sec:rq2}).} (a) Days between successive edits to a skill, per repository (488 intervals; box = IQR, whiskers 5--95\%, dashed line = overall median 5 days): cadence is frequent but organization-dependent, from a 1.5-day median (obra) to 14.5 (trailofbits). (b) Net change in a skill's resident token length between its first and last observation (120 skills with $\ge 2$ size observations): trajectories are growth-leaning, 32 growing more than 10\%, 7 shrinking more than 10\%, 81 stable, so the net-change distribution is predominantly stable-to-additive, with neither universal growth nor broad decay. Recomputed from the released records; the tail is clipped at $\pm 250\%$.}
\label{fig:rq2dyn}
\end{figure}

\section{Does Maintenance Help? The Transfer-Task Study}\label{app:rq4}
\textbf{Pre-registered design} (frozen before the run; full materials and harness released).
Skills: 13 public \texttt{SKILL.md} files with $\ge 6$ substantive edits, drawn from four repositories (the five deepest-history skills of the pilot plus eight more); \texttt{v\_old} = earliest in-window version at the current path (rename-resolved), \texttt{v\_new} = latest.
Tasks: 11 transfer tasks per skill (143 total), generated to exercise each skill's domain without telegraphing its conventions, then frozen and hashed; a curated minimal domain persona is given to the solver in every condition so the \texttt{no-skill} baseline stays clean.
Conditions \{\texttt{no-skill}, \texttt{v\_old}, \texttt{v\_new}\}, three independent replicates per cell (a per-replicate nonce makes the replicates genuine independent draws).
The solver is \texttt{gpt-5.4-mini}, a reasoning model from a different family than the judges, with truncation detection and retry that closes the earlier truncation confound.
The full skill package is exposed via an on-demand file-read tool (progressive disclosure), with \texttt{SKILL.md}-only and concatenated-package conditions retained as sensitivities.
An equalized multi-turn protocol (up to three turns) lets a constrained \texttt{gpt-4o-mini} user-simulator answer clarifying questions from the brief alone, so interaction-designed skills are not penalized; the judge rates the final output.

\textbf{Judging.} Each de-labelled, randomized triad is scored 1--5 on correctness, coverage, actionability, and non-redundancy by a sealed panel of blind judges, Claude models with no tool or web access, drawn from a different family than the solver to fix the pilot's same-provider judging; a \texttt{gpt-5.5} reference panel covers a 100-triad subset.
A scrub of namespaced sub-skill identifiers, skill slugs, version tokens, and organization names removes de-blinding fingerprints (a planted-fingerprint canary passes).
The inference unit is the skill ($n=13$); a pre-registered skill-level power simulation on paired skill-mean deltas gives 0.77 (sign test) to 0.90 ($t$-test) power to detect the pilot-sized effect, at a fixed $N$ (no optional stopping).

\textbf{Results.} The skill-level \texttt{v\_new}${}-{}$\texttt{v\_old} difference is $-0.09$ (95\% CI ${\approx}[-0.28,+0.10]$; sign-test $p=0.58$, Wilcoxon $p=0.55$; $d=-0.29$; 5 of 13 skills favor \texttt{v\_new}).
A length-controlled re-analysis (score residualized on log output length) gives $-0.06$ ($p=0.58$), so the null is not a length artifact; the deep-five subset agrees ($-0.14$).
No rubric dimension survives Holm correction, and cross-family judge agreement is moderate (Spearman $0.51$, below our pre-registered gate), so dimension-level patterns are exploratory and the aggregate null is conditional on a noisy but independent judging harness.
The gate was pre-registered as single-measures ICC(2,1)${}\ge{}$0.6, fixed at design freeze; computed on the 100-triad overlap it is $0.52$, below the gate and consistent with the upper bound that Pearson $0.57$ places on it.
A capability-bracket re-run of four of the deep-five skills with a weaker non-reasoning solver (\texttt{gpt-4o-mini}) yields a near-zero skill-level difference ($\approx 0$), so the null is not a ceiling effect of solver strength.
The exploratory comparison of the maintained version against the \texttt{no-skill} baseline is likewise null.
Table~\ref{tab:rq4} collects the battery.

\begin{table}[h]\centering\small
\begin{tabular}{@{}l l l >{\raggedright\arraybackslash}p{0.40\textwidth}@{}}
\toprule
Analysis & Unit & $\Delta$ (\texttt{v\_new}$-$\texttt{v\_old}) & Statistic / check \\
\midrule
Primary (full package) & 13 skills & $-0.09$ & sign $p{=}.58$, Wilcoxon $.55$; $d{=}{-}0.29$; CI $[-0.28,+0.10]$; 5/13 favor \texttt{v\_new} \\
Length-controlled & 13 skills & $-0.06$ & sign test $p{=}0.58$; score residualized on log output length \\
Deep-five subset & 5 skills & $-0.14$ & sign test $p{=}1.0$ \\
Weaker-solver bracket & 4 skills & $\approx +0.00$ & solver \texttt{gpt-4o-mini} (non-reasoning) \\
Harness reliability & 100 triads & $\rho{=}0.51$ & judge agreement: Pearson $0.57$; ICC(2,1)$\,{=}\,0.52$ (below $0.6$ gate) \\
\bottomrule
\end{tabular}
\caption{Robustness battery for the powered maintenance-benefit study. The primary skill-level test (top row) is the reportable result; the next three rows are sensitivity checks; the last row reports judging-harness reliability rather than a maintenance-benefit test. No rubric dimension survives Holm correction.}\label{tab:rq4}
\end{table}

\textbf{Limitations.} Tasks are author-and-model-constructed transfer scenarios rather than native repository workflows, and the second task author is a different \emph{model} (on a three-skill subset), a weaker control than an independent human writer.
The \texttt{no-skill} baseline is more verbose than the skilled conditions; the primary \texttt{v\_new}-versus-\texttt{v\_old} comparison is length-balanced, and the length-controlled re-analysis above absorbs any residual.
Every number regenerates from archived logs by the released analysis script; the harness, the frozen tasks, and per-skill provenance are released.

\subsection{Pilot study}\label{app:rq4-pilot}
The pilot covered the five skills with the deepest edit histories, with six transfer tasks per skill and two decoding seeds.
The solver, \texttt{gpt-4o-mini}, answered in a single turn and received only the \texttt{SKILL.md} file; the judge, \texttt{gpt-4o}, shared the solver's provider.
At the pre-registered 65\% effect size, the design had roughly 20\% statistical power.
Its two pre-registered tests pointed in different directions: a task-level sign test favored the maintained version at 15 wins to 6 with 9 ties ($p=0.08$), while a rubric-score Wilcoxon test reached $p=0.02$ with a mean gain of $+0.27$ points.
The comparison of the maintained version against the \texttt{no-skill} baseline was lopsided (12--45), an artifact of single-turn scoring: a skill that asks a clarifying question is penalized for an incomplete answer, which the powered study's equalized multi-turn protocol removes.
These limitations, in the package, the interaction protocol, the judge family, and the statistical power, motivated the powered study of Appendix~\ref{app:rq4}; under the corrected design, the pilot's positive trend does not reappear.

\section{A Replay Protocol for Skill Curation}\label{app:testbed}
The released corpus can be used for retrospective replay.
It includes trajectories and a scoring recipe, but no held-out leaderboard and no baseline curator; for that reason we describe it as a protocol rather than a benchmark.
An automated skill curator is evaluated by replaying historical maintenance:
\begin{itemize}\itemsep1pt
\item \textbf{Input}: a prior \texttt{SKILL.md} version plus the context fields available at that point (commit message, linked issue/PR, repository).
\item \textbf{Target}: the next human substantive edit, scored by the operation label, the component touched, and similarity to the actual patch or summary.
\item \textbf{Split}: chronological per skill (predict each edit from earlier history) or repository-held-out (train on four repositories, test on the fifth), with the split named by the released pinned commits.
\item \textbf{Metrics}: operation and component match rates; normalized token overlap with the actual diff; ROUGE-L for one-sentence summaries; and a pruning check. The pruning check asks whether the curator ever consolidates or deprecates content rather than only adding it. The observed human rate is about 4\%, but it is noisy and selection-dependent; use it as a reference, not a target.
For content edits, the diff can also be scored by \emph{atomic-fact coverage}, following WiNELL \citep{reddyWiNELL2026}.
Decompose the human edit into atomic units, then credit a \emph{soft} match if the fact appears anywhere in the file and a \emph{hard} match if it appears in the correct component.
WiNELL finds that models more often identify \emph{what} to add than \emph{where} to put it.
That is the placement problem isolated by our component analysis (\S\ref{sec:rq2}, Appendix~\ref{app:component}).
\end{itemize}
\textbf{Scope.} The context fields may reveal the edit's rationale after the fact, so replay is a \emph{hindsight} task.
A low match rate is still informative: the curator cannot reconstruct edits humans made even when given the historical context around the edit.
A high match rate is weaker evidence.
It does not show that the same curator could have discovered the edit prospectively, and matching human edits is not deployable autonomous curation.
Running a baseline on this protocol is future work.

\end{document}